\documentclass{article}

\usepackage{microtype}
\usepackage{graphicx}
\usepackage{subcaption}
\usepackage{booktabs} % for professional tables
\usepackage{lingmacros}
\usepackage{tree-dvips}
\usepackage{multicol}

\usepackage{hyperref}

\usepackage[accepted]{icml2019}

\icmltitlerunning{A Deep Learning-based Framework for the Detection of Schools of Herring in Echograms}

\begin{document}

\twocolumn[
\icmltitle{A Deep Learning-based Framework for the Detection \\ 
of Schools of Herring in Echograms}



\icmlsetsymbol{equal}{*}

\begin{icmlauthorlist}
\icmlauthor{Alireza Rezvanifar}{equal,Uvic}
\icmlauthor{Tunai Porto Marques}{equal,Uvic}
\icmlauthor{Melissa Cote}{Uvic}
\icmlauthor{Alexandra Branzan Albu}{Uvic}
\icmlauthor{Alex Slonimer}{ASL}
\icmlauthor{Thomas Tolhurst}{ASL}
\icmlauthor{Kaan Ersahin}{ASL}
\icmlauthor{Todd Mudge}{ASL}
\icmlauthor{St\'ephane Gauthier}{DFO}
\end{icmlauthorlist}

\icmlaffiliation{Uvic}{University of Victoria, Victoria, Canada}
\icmlaffiliation{ASL}{ASL Environmental Sciences, Victoria, Canada}
\icmlaffiliation{DFO}{Fisheries and Oceans Canada, Victoria, Canada}

\icmlcorrespondingauthor{Alexandra Branzan Albu}{aalbu@uvic.ca}
% \icmlcorrespondingauthor{Eee Pppp}{ep@eden.co.uk}

\icmlkeywords{Machine Learning, Deep Learning, Echograms, Fish Dectection}

\vskip 0.3in
]



\printAffiliationsAndNotice{\icmlEqualContribution} % otherwise use the standard text.

\begin{abstract}
Tracking the abundance of underwater species is crucial for understanding the effects of climate change on marine ecosystems. Biologists typically monitor underwater sites with echosounders and visualize data as 2D images (echograms); they interpret these data manually or semi-automatically, which is time-consuming and prone to inconsistencies. This paper proposes a deep learning framework for the automatic detection of schools of herring from echograms. Experiments demonstrated that our approach outperforms a traditional machine learning algorithm using hand-crafted features. Our framework could easily be expanded to detect more species of interest to sustainable fisheries.
\end{abstract}

\begin{figure*}[t!]
\centering
\includegraphics[scale=0.45]{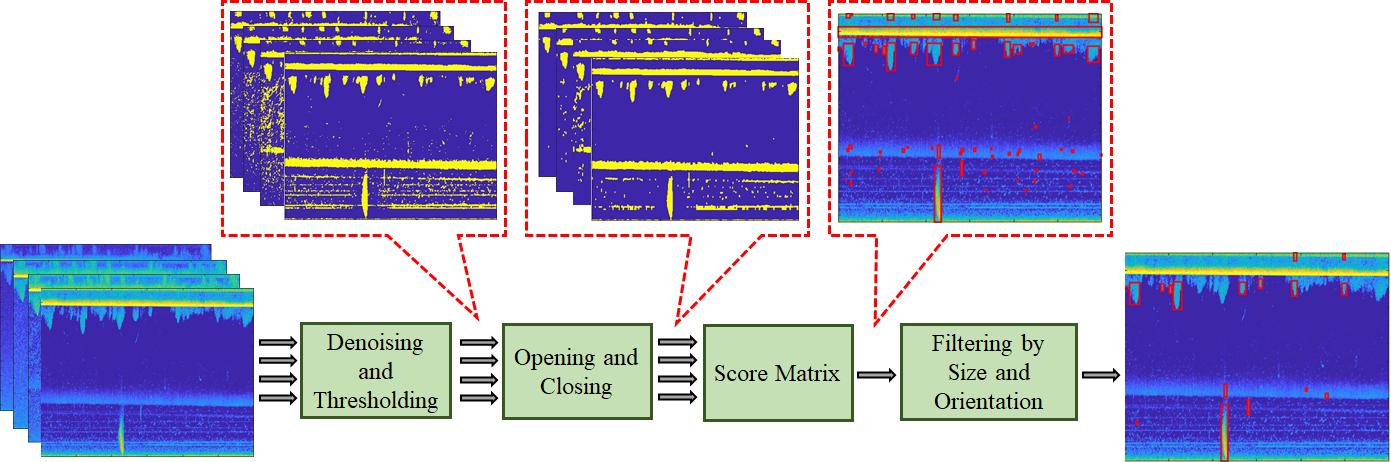}
\vspace{-1.2em}
\caption{ROI extractor pipeline and sample processed echograms at each stage.}
\vspace{-4mm}
\label{ROIext}   % Give a unique label
\end{figure*}

\section{Introduction}
\label{intro}
The study of acoustic backscatter is commonly utilized by biologists as a means to monitor underwater sites in a thorough and non-invasive manner. Different species possess different acoustic properties, thus producing different responses. Acoustic survey data are typically interpreted via time-consuming manual or semi-automatic approaches prone to inconsistencies, using expensive softwares such as Echoview (https://www.echoview.com/). In this paper, we apply deep learning methods for automatically detecting specific biological targets in acoustic survey data. Our goals are to improve data processing and interpretation for species abundance tracking and environmental monitoring, and to provide a tool that reliably supports analyses of the effects of climate change. We present a case study on schools of herring.

Acoustic data are obtained via multifrequency echosounders, such as the Acoustic Zooplankton Fish Profiler (AZFP) \cite{lemon2012multiple}, and visualized in 2D, with depth along the y-axis and time along the x-axis, producing an echogram. A color code represents the echo amplitude, usually in the form of volumetric backscatter strength. Visual structures indicate geological and biological elements. Acoustic-scattering models provide a link between echograms and the size/abundance distribution of the animals \cite{trevorrow2005use}. Echograms are traditionally analyzed using statistical characteristics of the aggregations of organisms \cite{stanton201230}. Feature-based classification methods commonly favor a classical machine learning \cite{lefeuvre2000acoustic,cabreira2009artificial,charef2010classification,robotham2010acoustic,gauthier2014species,fallon2016classification} paradigm and use hand-crafted features extracted with Echoview. Deep learning, which has been shown to be very effective at various tasks in computer vision such as object detection and recognition, has yet to permeate echogram analysis. Only a few methods deal with identifying fish from echograms utilizing deep learning and are quite limited in the nature of their experiments \cite{hirama2017discriminating} or data \cite{shang2018study}. A few additional works have applied deep learning to the detection of mobile marine life in sonar data \cite{liu2018counting,french2018jellymonitor,glukhov2019joint}, which typically have a higher resolution than echograms.

Our contributions are two-fold. Firstly, we propose a novel application of deep learning techniques for fish detection from echograms that goes beyond the few existing works. Secondly, our framework automates acoustic survey analyses; an automatic approach will reduce processing times, required man-power, and inconsistencies in the results. Our approach also has the potential to be scaled to handle additional underwater species. Biologists collect large quantities of echograms so the amount of data is not a limitation for the training of a deep learning-based framework; however, the annotation process is a tedious task. We show that, even using a smaller annotated dataset, deep learning-based solutions can outperform a classical machine learning method. Our promising results indicate that with the annotation of more data, deep learning can be efficiently applied in underwater acoustic analysis. 

\section{Technical Approach}
\label{ourwork}
Thanks to the emergence of convolutional neural networks (CNNs) and the existence of large annotated datasets like MS COCO \cite{lin2014microsoft}, PASCAL VOC \cite{everingham2010pascal}, and ImageNet \cite{krizhevsky2012imagenet}, object detection is now well addressed for natural scenes. CNNs have been utilizing different architectures to find regions of interest (ROIs) in images and to classify them. While earlier networks like R-CNN \cite{girshick2014rich} and Fast R-CNN \cite{girshick2015fast} were using selective search \cite{uijlings2013selective} for finding ROIs, newer ones like Faster R-CNN\cite{ren2015faster} use another network (region proposal network (RPN)). Unfortunately, this approach is not feasible in applications outside the natural scene scope, including ours, where there is insufficient box-annotated data for training RPNs. We thus designed an ROI extractor based on high level information on the intensity and morphology of schools of herring. We introduce the proposed ROI extractor in Section \ref{ROI} and then describe the deep learning-based classification phase in Section \ref{sub-section:classification}. Our data are described next in Section \ref{dataset}.

\subsection{Dataset}
\label{dataset}
Our dataset is composed of 100 multifrequency echograms generated from data collected with an AZFP device deployed on the surface of the water looking downward, in the Discovery Passage off Vancouver Island, British Columbia, Canada in 2015. The AZFP transducer was calibrated in four frequency channels (67, 125, 200, and 455 KHz) to detect schools of herring more efficiently. We use all four frequencies as input, as they may convey complementary information. The echogram resolution is 1200 x 571 pixels, representing one hour of measurements in a depth range of 50 meters. The dataset is divided into 70 echograms for training (which itself is divided in 80\% training, 20\% validation) and 30 echograms for testing.

\begin{figure*}[t!]
  \includegraphics[width=0.331\textwidth]{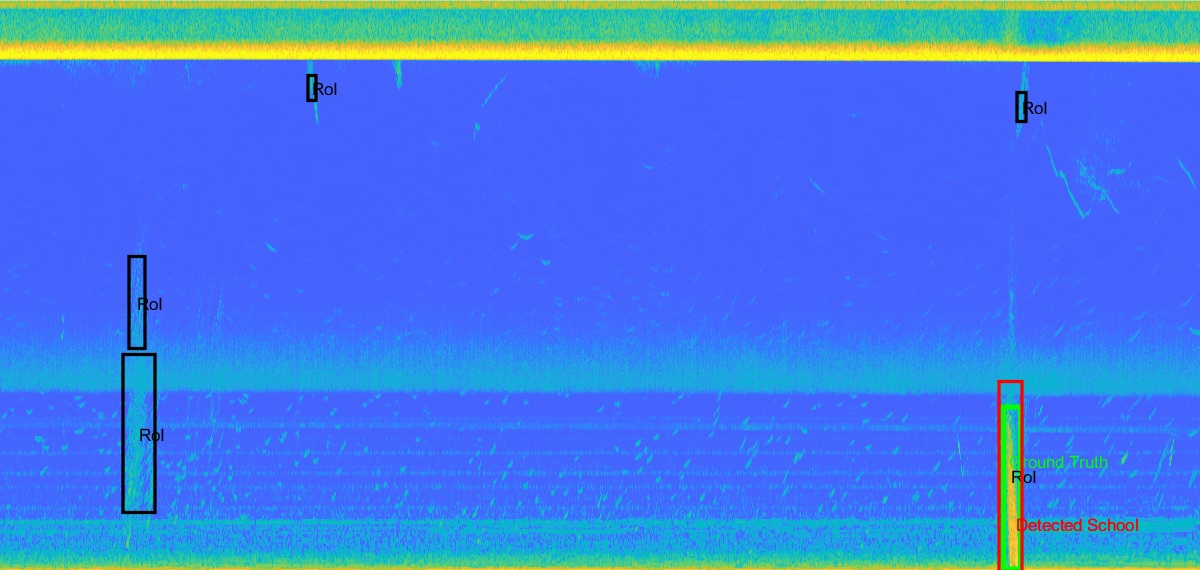}
  \includegraphics[width=0.331\textwidth]{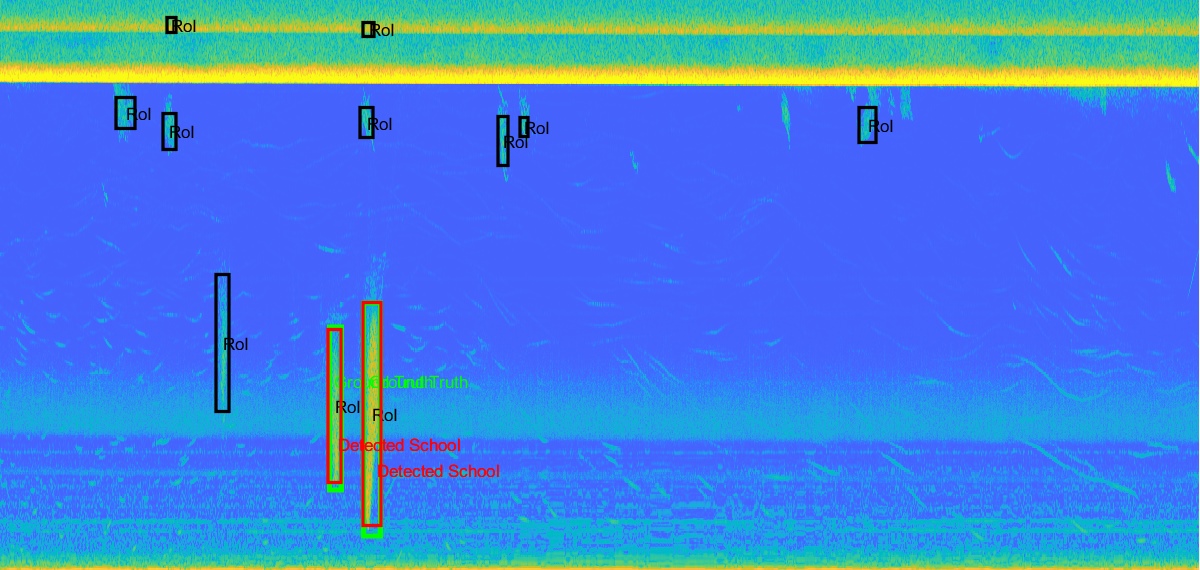}
  \includegraphics[width=0.331\textwidth]{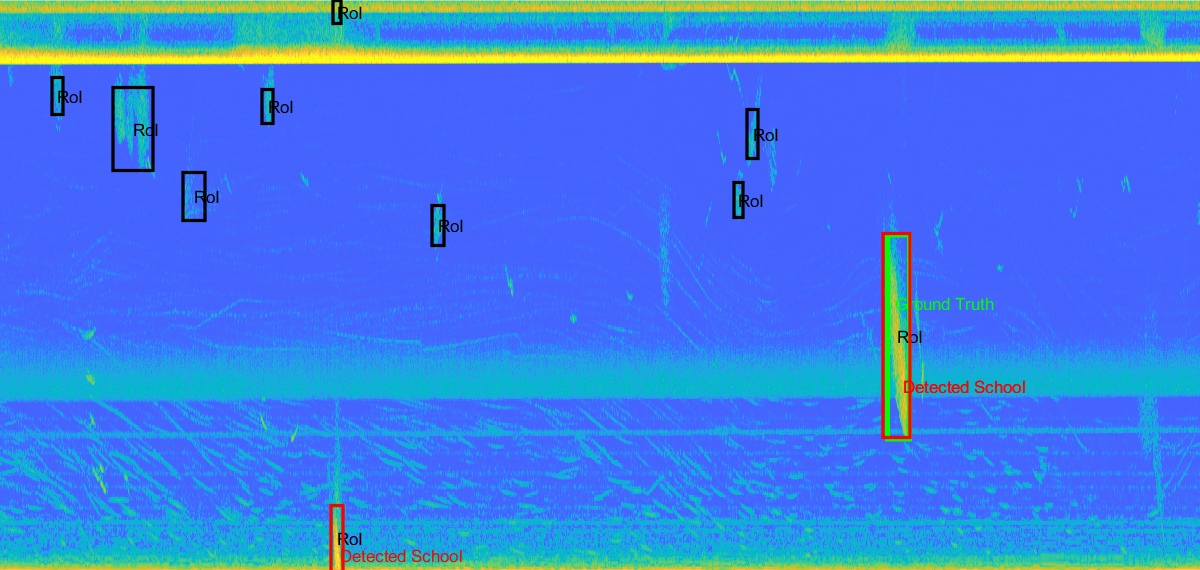}
  \vspace{-2.4em}
  \caption{Results of the detection framework using DenseNet. Black: regions classified as background; green: ground truth bounding boxes; red: regions classified as schools of herring. All detections are correct in the first two samples (left and center), while one ROI bounding box was incorrectly classified in the right-most echogram.}
\vspace{-4mm}
\label{pipeline_results}
\end{figure*}

\subsection{ROI Extractor}
\label{ROI}
Two high level features common to all schools of herring are presumed: 1) a strong intensity core visible in most echograms, 2) a vertical elongated shape. Figure \ref{ROIext} shows the ROI extractor pipeline and sample intermediate outputs after each step. We first apply a median filter along the time direction of the echograms to smooth out the spiky variations and remove some of the undesirable signal such as small fish. The denoised echograms are then binarized using adaptive thresholding \cite{bradley2007adaptive}, and processed via morphological opening and closing to remove small isthmuses and protrusions and fill in small holes. We create a binary score matrix, indicating the strength of the potential ROIs, by summing the four processed binarized echograms and retaining the pixels that have a value greater than 2, thus retrieving objects present in at least three channels. Connected components are then filtered for size and orientation: we discard very small regions (less than 50 pixels) as well as regions with an orientation smaller than 60 degrees with respect to the horizontal axis, as those cannot be schools of herring. The remaining connected components constitute the final ROIs. The ROI bounding boxes are passed to the classification stage (Section \ref{sub-section:classification}) for further analysis.

We do not seek a high precision for ROIs at this stage, as they will be further refined in the next step. However, we do seek a high recall; if a school of herring is not included as an ROI, it cannot be recovered in subsequent steps.

\subsection{Classification}
\label{sub-section:classification}
The classification stage determines whether each ROI represents a school of herring or background. We compare three deep learning-based architectures, which automatically extract features (without any contextual information) from ROIs and classify them: ResNet \cite{he2016deep}, DenseNet \cite{huang2017densely} and InceptionV3 \cite{szegedy2016rethinking}.

Each echogram might contain multiple instances of schools of herring (\textit{positive} samples) and background (\textit{negative} samples). To obtain meaningful and realistic negative samples (avoiding the use of random crops), we used the ROI extractor outputs for the 70 training echograms as follows. The ROI bounding boxes are compared with the positive samples from the ground truth and the intersection-over-union (IoU), i.e.\@ ratio between areas of overlap and union, is calculated. ROI bounding boxes with less than $0.4$ of IoU score are used as negative samples and the others as positive samples. We follow the 1:2 ratio between positive and negative samples used in the popular work of Girshick et al. \cite{girshick2015fast}. 

\section{Results and Discussion}
\label{results}

The two main components of the framework (ROI extractor and classifier) are evaluated separately using precision, recall, and F1-score metrics.

\subsection{ROI Extractor Evaluation}
\label{sub-section:ROI_evaluation}
Since we are dealing with bounding boxes, we make use of an IoU threshold between the ROIs and the ground truth to determine true positives, false positives, and false negatives. We consider a true positive when an ROI bounding box has an IoU with the ground truth \textit{higher} than the threshold. Results for the 100 echograms are shown in Table \ref{tab:ROI_eval}.

\begin{table}[h]
\vspace{-4mm}
\small
\caption{Precision, recall and F1-score of the ROI extractor for different IoU thresholds.}
\label{tab:ROI_eval}
\centering

\begin{tabular}{c c c c}
\toprule
\textbf{IoU Threshold}&\textbf{Precision}&\textbf{Recall}&\textbf{F1-Score}\\
\midrule
0.0 & 0.173 & \textbf{0.931} & 0.292 \\
0.2 & 0.171 & 0.917 & 0.288 \\
0.4 & 0.155 & 0.834 & 0.262 \\
\bottomrule
\end{tabular}
\vspace{-3.5mm}
\end{table}

The most relevant metric here is the \textit{recall}, as ROIs are further classified in the next stage and the classifier cannot recover a school missed by the ROI extractor. Table \ref{tab:ROI_eval} shows that the performance increases as the IoU threshold is lowered. The small difference in recall (less than 10 p.p.) between IoU thresholds 0.0 and 0.4 shows that most of the ROI bounding boxes have a significant overlap with the ground truth. Ideally, the incorrect bounding boxes will be identified as background in the next step, classification.  

\subsection{Framework Evaluation}
\label{sub-section:framework_evaluation}

This section assesses the complete framework as the output of the ROI extractor are used as input to the classifiers. Table \ref{tab:pipeline_eval} shows the classification results for each tested network architecture over the test set (30 echograms) for an IoU threshold of 0.4, while highlighting the best results in bold. We compare our results with a baseline method based on traditional hand-crafted features engineered for fisheries research (mean intensity, eccentricity, and circularity of the region) and classical machine learning (support vector machine (SVM) classifier \cite{cortes1995support} with a linear kernel).

The DenseNet architecture achieves the best balance between metrics, and outperforms the baseline method. Figure \ref{pipeline_results} shows sample results of the framework using DenseNet. In most of the echograms (as examplified by the left and center images of Figure \ref{pipeline_results}), the ROI extractor was able to detect bounding boxes around the positive samples, followed by a correct classification by the DenseNet. However, the classifier also incorrectly classified some samples, as illustrated in the right-most image of Figure \ref{tab:pipeline_eval}. Experiments where the IoU threshold was set to 0.0 in the classification phase saw a recall by the framework that reached the maximum value of 0.93 (bounded by the ROI extractor), attesting to the efficiency of the trained classifier. However, the usage of such a low IoU threshold is not reasonable, given that any ROI bounding box that at least touches a positive sample is considered as a school of herring candidate. Higher IoU thresholds guarantee that the detections are more reliable.

\begin{table}[H]
\vspace{-4mm}
\small
\caption{Precision, recall and F1-score of the full pipeline using different CNN-based architectures for IoU thresholds of 0.4.}
\label{tab:pipeline_eval}
\centering
\begin{tabular}{c c c c}
\toprule
\textbf{Architecture}&\textbf{Precision}&\textbf{Recall}&\textbf{F1-Score}\\
\midrule
ResNet50 \cite{he2016deep} & 0.77 & \textbf{0.85} & 0.81\\
DenseNet201 \cite{huang2017densely} & 0.78 & \textbf{0.85} & \textbf{0.82}\\
InceptionNet \cite{szegedy2016rethinking} & \textbf{0.81} & 0.81 & 0.81\\
Baseline (SVM) \cite{cortes1995support} & 0.51 & 0.78 & 0.62 \\
\bottomrule
\end{tabular}
\vspace{-4.1mm}
\end{table}

\section{Conclusion}
\label{conclusion}
We propose a novel framework that assists the study of species' abundance in underwater sites using acoustic data. The first case study of the framework is focused on the detection of schools of herring in echograms using two main components: a novel ROI extractor and a deep learning-based image classifier. The framework achieved a good performance for intersection-over-union thresholds (i.e.\@ 0.4) which guaranteed meaningful detection. This framework will assist in the interpretation of large amounts of raw acoustic data. Although this proof-of-concept focuses on schools of herring, it is scalable to other aquatic species that can be monitored using echosounders (e.g.\@ zooplankton, salmon). The future substitution of our ROI extractor by generic ROI calculation systems (e.g. region proposal networks \cite{ren2015faster}) will simplify the expansion of detectable classes to the gathering of new visual samples. The ability to measure the abundance of such subjects over extended periods of time constitutes a strong tool for the study of the effects of water temperature shifts caused by climate change-related phenomena.

%% Acknowledgements should only appear in the accepted version.
%\section*{Acknowledgements}

\bibliography{SHdetection}
\bibliographystyle{unsrt}

%\appendix
%\section{Do \emph{not} have an appendix here}
%
%\textbf{\emph{Do not put content after the references.}}
%
%Put anything that you might normally include after the references in a separate
%supplementary file.
%
%We recommend that you build supplementary material in a separate document.
%If you must create one PDF and cut it up, please be careful to use a tool that
%doesn't alter the margins, and that doesn't aggressively rewrite the PDF file.
%pdftk usually works fine. 
%
%\textbf{Please do not use Apple's preview to cut off supplementary material.} %In
%previous years it has altered margins, and created headaches at the %camera-ready
%stage. 
%%%%%%%%%%%%%%%%%%%%%%%%%%%%%%%%%%%%%%%%%%%%%%%%%%%%%%%%%%%%%%%%%%%%%%%%%%%%%%%
%%%%%%%%%%%%%%%%%%%%%%%%%%%%%%%%%%%%%%%%%%%%%%%%%%%%%%%%%%%%%%%%%%%%%%%%%%%%%%%

\end{document}